# Deep Learning-based Detection of Bacterial Swarm Motion Using a Single Image

Yuzhu Li[†,1,2,3], Hao Li[†,4], Weijie Chen[†,4,5], Keelan O'Riordan[1,6], Neha Mani[7], Yuxuan Qi[1,8], Tairan Liu[1,2,3], Sridhar Mani[*,4], and Aydogan Ozcan[*,1,2,3,9]

[1]Electrical and Computer Engineering Department, University of California, Los Angeles, CA, 90095, USA.

[2]Bioengineering Department, University of California, Los Angeles, 90095, USA.

[3]California NanoSystems Institute (CNSI), University of California, Los Angeles, CA, 90095, USA.

[4]Department of Medicine, Genetics and Molecular Pharmacology, Albert Einstein College of Medicine, Bronx, NY, 10461, USA.

[5]Intelligent Medicine Institute, Shanghai Medical College, Fudan University, Shanghai, 200032, China.

[6]Department of Mathematics, University of California, Los Angeles, CA, 90095, USA.

[7]Columbia University, New York, NY, 10027, USA.

[8]Department of Computer Science, University of California, Los Angeles, CA, 90095, USA.

[9]Department of Surgery, University of California, Los Angeles, CA, 90095, USA.

**\*Correspondence: Aydogan Ozcan,** ozcan@ucla.edu **Sridhar Mani,** sridhar.mani@einsteinmed.edu

[†] **Equal contributing authors**

## Abstract

Motility is a fundamental characteristic of bacteria for nutrient acquisition, and evasion of hostile environments. Distinguishing between swarming and swimming, the two principal forms of bacterial movement, holds significant conceptual and clinical relevance. This is because bacteria that exhibit swarming capabilities often possess unique properties crucial to the pathogenesis of infectious diseases and may also have therapeutic potential. Conventionally, the detection of bacterial swarming involves inoculating samples on an agar surface and observing colony expansion, which is inherently qualitative, time-intensive, and typically requires additional testing to rule out other forms of motility. A recent methodology that differentiates swarming and swimming motility in bacteria using circular confinement offers a rapid approach to detecting swarming. However, it still heavily depends on the observer's expertise, making the process labor-intensive, costly, slow, and susceptible to inevitable human bias. To address these limitations, we developed a deep learning-based swarming classifier that rapidly and autonomously predicts swarming probability using a single blurry image. Compared with traditional video-based, manually-processed approaches, our method is particularly suited for high-throughput environments and provides objective, quantitative assessments of swarming probability. The swarming classifier demonstrated in our work was trained on *Enterobacter sp.* SM3 and showed good performance when blindly tested on new swarming (positive) and swimming (negative) test images of SM3, achieving a sensitivity of 97.44% and a specificity of 100%. Furthermore, this classifier demonstrated robust external generalization capabilities when applied to unseen bacterial species, such as *Serratia marcescens* DB10 and *Citrobacter koseri* H6. It blindly achieved a sensitivity of 97.92% and a specificity of 96.77% for DB10, and a sensitivity of 100% and a specificity of 97.22% for H6. This competitive performance indicates the potential to adapt our approach for diagnostic applications through portable devices or even smartphones. This adaptation would facilitate rapid, objective, on-site screening for bacterial swarming motility, potentially enhancing the early detection and treatment assessment of various diseases, including inflammatory bowel diseases (IBD) and urinary tract infections (UTI).

## Introduction

Motility is an intrinsic characteristic of bacteria; despite the energy expenditure, it provides high returns by enabling bacteria to acquire nutrients actively and evade harmful environments[1]. Swimming/planktonic and swarming are the two primary forms of bacterial movement. Swimming involves individual bacteria propelling themselves in a single direction using flagella in liquid environments. Swarming, however, entails rapid coordinated movement of bacterial groups in the same direction on semi-solid surfaces, facilitated by flagella and surfactants[2]. These two concepts, swarming and swimming, both mediated by flagella, are sometimes conflated; for example, densely populated swimming bacteria are often referred to as "a swarm of bacteria." However, most microbiologists recognize that swarming and swimming represent fundamentally different types of motilities. Understanding their distinctions holds significant research and clinical value, because swarming bacteria can exhibit unique properties that are absent in bacteria lacking swarming capabilities. Swarming bacteria can move effectively on semi-solid surfaces, and they exhibit distinctive properties when infecting hosts or demonstrating therapeutic effects, compared to bacteria that possess only swimming abilities without swarming capabilities. For example, in patients with catheter and non-catheter-associated urinary tract infections (UTI), 80% of the pathogenic clinical isolates are *uropathogenic E.coli* (UPEC)[3]. The primary determinant of UPEC virulence in UTI is swarm (biofilm) formation[4,5]. In mice, UPEC moves up from the bladder to infect the kidney; however, swarm-deficient UPEC is unable to infect the kidney[6]. Another example is *Proteus sp.* (*P. mirabilis*), a swarming bacteria whose urinary tract virulence is notably associated with swarming behavior[7–10]. Contrasting with the generally harmful effects typically related to swarming bacteria, recent studies suggest that the enrichment of swarming bacteria under conditions of intestinal stress may confer beneficial effects. Specifically, swarming bacteria have been shown to promote intestinal mucosal repair and alleviate inflammation, thus demonstrating therapeutic effects for inflammatory bowel diseases (IBD)[11]. Consequently, swarming bacteria may serve as effective biomarkers for detecting several diseases. Given these findings, there is a pronounced need to develop efficient and cost-effective methods for detecting the presence of swarming bacteria in e.g., urine and fecal samples, thereby facilitating the diagnosis of infections or related diseases such as IBD[12].

For a conventional bacterial motility test, researchers inoculate the sample on a 0.5% (w/v) or 0.3% agar surface and observe the colony expansion to determine if the sample bacteria are swimming or swarming. However, this qualitative method requires at least 10 hours, and additional experiments are needed to exclude gliding or sliding motility. To this end, prior work has successfully distinguished the motion patterns of planktonic and swarming *Enterobacter sp.* SM3 with high efficiency and certainty on the soft agar surface, which was facilitated by the use of Polydimethylsiloxane (PDMS) chips with circular wells[13] to amplify the differences between swarming and swimming motion patterns for clearer observation[14,15]. This prior research has investigated the visual and physical differences between SM3 swarmers and SM3 swimmers in confinements using naked-eye based observations assisted by Vortex Order Parameters (VOP) and spatial auto-correlation functions[13]. Specifically, in addition to direct visual checks, image sequences of each captured video were used to calculate the velocity field and then calculate the parameters and functions[16], revealing that swarmers exhibit a single-swirl motion pattern whereas their planktonic counterparts form multiple randomly-distributed local swirls within the confinements. This process requires manually marking the region of interest (ROI) and smoothing the vector field, which is relatively laborious, time-consuming, and prone to errors and thus not suitable for high-volume and automated settings. Moreover, the determination of the VOP thresholds for differentiating swarming and swimming motion patterns is subjective and relies heavily on experiential judgment, making the ambiguous patterns hard to define and further complicating the interpretation of results. Thus, there is an urgent need for an automatic, objective, and quantitative image recognition method, capable of making accurate and rapid decisions on bacterial motility types.

The rapid advancement of artificial intelligence (AI) in recent years offers a promising avenue to address this problem. Deep neural networks (DNNs) stand out due to their exceptional ability to handle high-dimensional, sparse, and noisy data exhibiting nonlinear relationships[17]. This capability allows DNNs to identify subtle distinctions between similar images or videos, even in varied and noisy environments. Consequently, DNNs have been widely adopted across a broad spectrum of biomedical fields, such as microorganism detection[18–23], disease detection[24–28], and cell[29,30], organelle[31] and organ segmentation[32], among others.

Here, we report a novel method of determining the type of bacterial motility on a surface. Unlike traditional video-based human-intervened swarming detection methods (Fig. 1(a)), this deep learning-empowered method can rapidly and automatically identify swarming events from non-swarming planktonic motion patterns using a single blurry image, as shown in Fig. 1(b). The swarming DNN classifier, developed using an attention-based neural network, was trained to distinguish between swarming and planktonic motion patterns by interpreting specific spatiotemporal features encoded in a single image through a long integration time, yielding high specificity and sensitivity for both internal generalization (SM3 bacteria) and external generalization tests involving other types of bacteria never seen before, such as *Serratia marcescens* DB10 and *Citrobacter koseri* H6. This technique offers an automated, objective, efficient, and quantitative approach for evaluating the swarming probability of a sample. This advancement has the potential to evolve into a portable system or be integrated into a smartphone-based device together with simple disposable chips, creating a convenient, rapid, and practical method for on-site screening of microbial motility in complex samples and helping the specific and sensitive detection of various bacterial diseases.

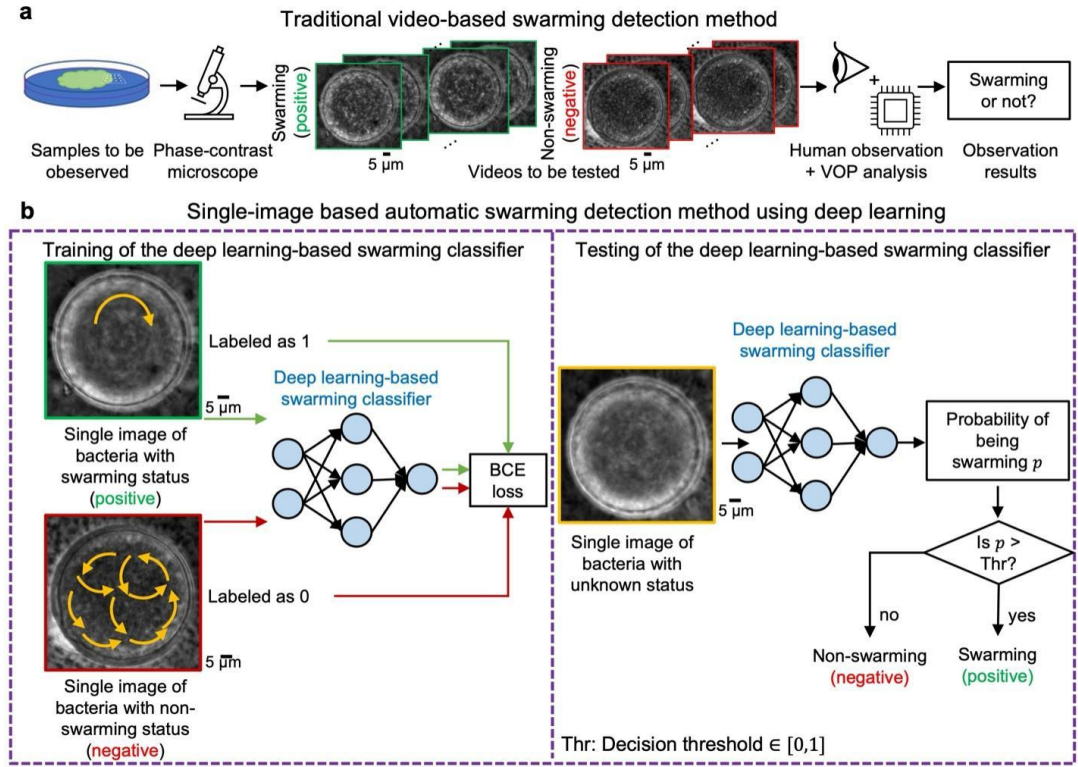

**Figure 1. Deep learning-enabled bacterial motility analysis to detect swarming motion on a surface. (a)** The traditional

video-based swarming detection methods require human interventions and observations, which are time-consuming, laborious, and susceptible to various sources of errors. **(b)** The deep learning-based swarming classifier, after its training, can rapidly provide automated, and accurate predictions of bacterial motion patterns (swarming or planktonic) on unseen data using a single image with a long integration time of ~0.3 sec.

## Results

The deep learning-enabled bacterial swarming detection neural network used in our work was built based on the DenseNet[33] architecture using SM3 (*Enterobacter sp.*) bacteria. Samples exhibiting swarming and planktonic states of SM3 were prepared according to a standardized protocol outlined in ref[13] and detailed in the Methods section. These samples were then positioned under a phase-contrast microscope (Olympus CKX41, 20×) for video capture. The single image per well, serving as the input of the deep learning-based swarming classifier, was obtained by averaging 10 consecutive frames from the raw video, corresponding to a total integration time of ~0.34 sec per image. In this single blurry image, the time-varying bacterial movement is encoded into the spatial domain, creating distinct trajectory patterns for the swarming and swimming states of bacteria that the DNN model can learn and differentiate. Specifically, swarming (positive) patterns are characterized by a distinct, centrally located bright ring in each long exposure image, arising from the collective and repetitive global swirling motion of SM3 swarmers. In contrast, swimming (negative) patterns resulting from the free movement of individual bacteria produce more random and featureless trajectory patterns in each long exposure image. However, a bright ring-shaped artifact located at the well edges was also observed in both the swarming and swimming bacteria samples, complicating the predictions made by the DNN model. This occurs because the bacterial movement is confined at the well edges, forcing the bacteria to move along these boundaries. To address this edge artifact and enhance the model accuracy, an attention module was integrated to the basic structure of the classifier (as shown in Fig. 2). This module adaptively adjusts the radius and centroid shifts of the circular active region for each well to mitigate the influence of peripheral bright rings.

The deep learning-based swarming classifier was trained using a dataset of 1,301 positive images from 52 wells and 2,703 negative images from 38 wells, with 90% allocated for training and the remaining 10% reserved for validation. The trained network model would predict a continuous output value ranging from 0 to 1 for each unseen test image, representing the likelihood of being swarming motility. Using a predefined decision threshold, the model blindly categorized each test well as either swarming or non-swarming based on its prediction score.

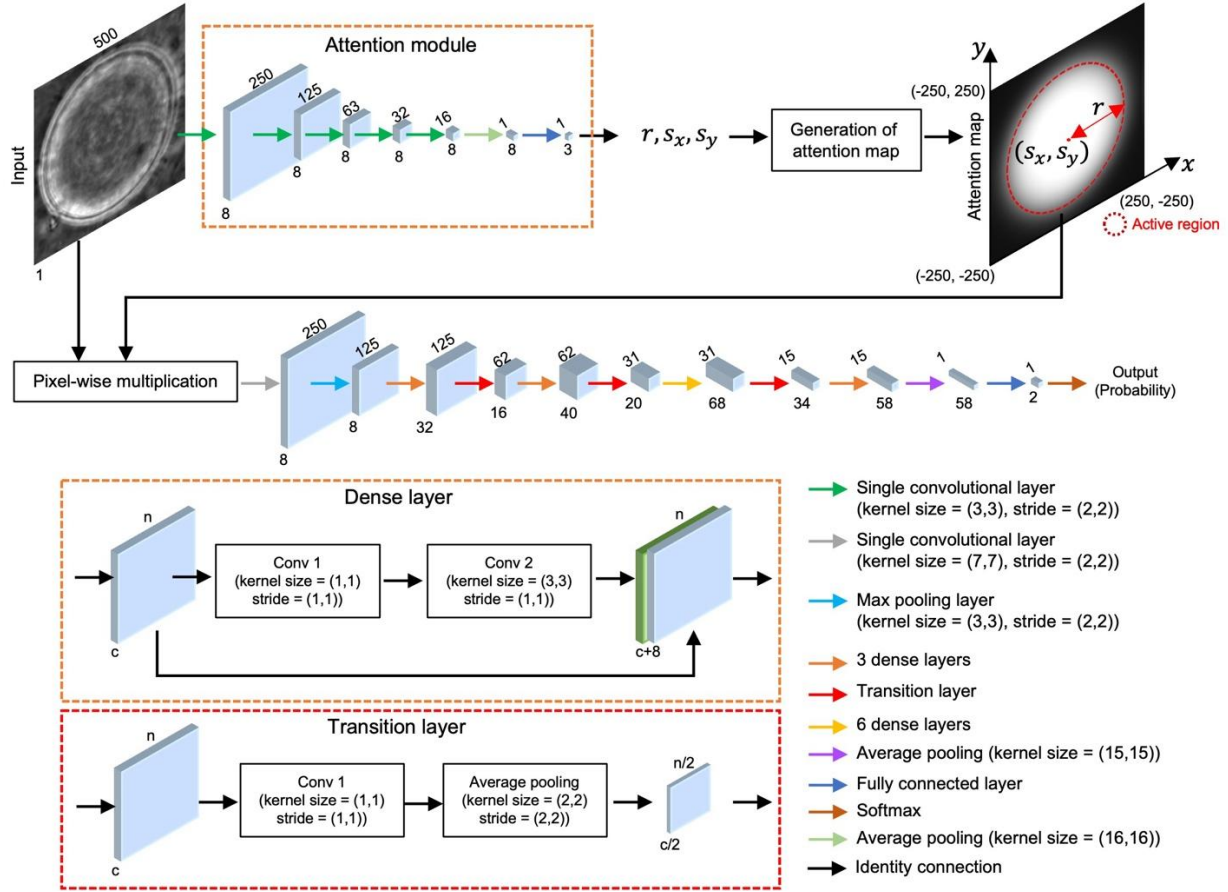

**Figure 2. Network architecture of the deep learning-based swarming classifier.** The DNN model was built based on the backbone of DenseNet[33]. The attention module is used to adaptively generate attention maps with adjustable radius and centroid shifts of the circular active region for each well.

To blindly test the efficacy of our deep learning-based swarming classifier, we initially conducted tests on its internal generalization capabilities using new, unseen SM3 samples (the same type of bacteria used in training). Figure 3(a) showcases three representative images for each of the SM3 swarming and swimming/planktonic patterns, along with the swarming probabilities blindly predicted by our DNN model. The predicted swarming probabilities closely aligned with the ground truth labels, assigning high values to swarming patterns and low values to swimming patterns, demonstrating the model's accuracy in distinguishing between these motility patterns. Subsequently, the swarming detection DNN model underwent blind testing on a large dataset comprising 39 SM3 swarming (positive) well images and 44 SM3 swimming/planktonic (negative) well images, achieving a sensitivity of 97.44% and a specificity of 100%. Detailed confusion matrix results are presented in Fig. 3(a).

Next, we extended our blind testing of the deep learning-based swarming classifier to include two new bacterial strains, DB10 (*Serratia marcescens* Lab strain) and H6 (*Citrobacter koseri*), which were not included in the training set; this constitutes an external generalization test on the trained classifier. The classification results are summarized in Fig. 3(b). For DB10, when tested on 48 swarming (positive) well images and 31 swimming/planktonic (negative) well images, our DNN model achieved a sensitivity of 97.92% and a specificity of 96.77%. Similarly, for H6, the model was tested on 27 swarming (positive) well images and 36 swimming/planktonic (negative) well images, achieving a sensitivity of 100% and a specificity of 97.22%. Detailed confusion matrices for the DB10 and H6 classification performance are also presented in Fig. 3(b). The high accuracy achieved with these new bacterial types demonstrates the robust external generalization capability of the trained DNN model. This suggests that the features the network learned to differentiate between swarming and swimming patterns are not specific to a type of bacteria, which allows for successful generalization to other bacteria strains without the need for retraining or transfer learning, highlighting the model's versatility and adaptability. Three exemplar positive videos (one each for SM3, DB10, and H6) and three exemplar negative videos (one each for SM3, DB10, and H6)) together with their corresponding swarming probability scores predicted by our deep learning-based swarming classifier are reported in Supplementary Videos 1 and 2, respectively.

To further evaluate the classification performance of our deep learning-based swarming classifier, we plotted the changes in sensitivity and specificity across varying decision thresholds for the SM3, DB10, and H6 strains, as shown in Fig. 4(a, d, g) for sensitivity and Fig. 4(b, e, h) for specificity. Additionally, Receiver Operating Characteristic (ROC) curves for each bacteria type are depicted in Fig. 4(c, f, i), illustrating the trade-off between the true positive and the false positive rates at different threshold settings. The Area Under the Curve (AUC) is also reported in each case, demonstrating a very good classifier performance with scores of 0.9988 for SM3, 0.9933 for DB10, and a perfect score of 1.0 for H6. These metrics underscore the swarming classifier's high accuracy and robustness, demonstrating its strong internal and external generalization performance, also highlighting its effectiveness in accurately distinguishing between swarming and planktonic motion patterns across different types of bacteria.

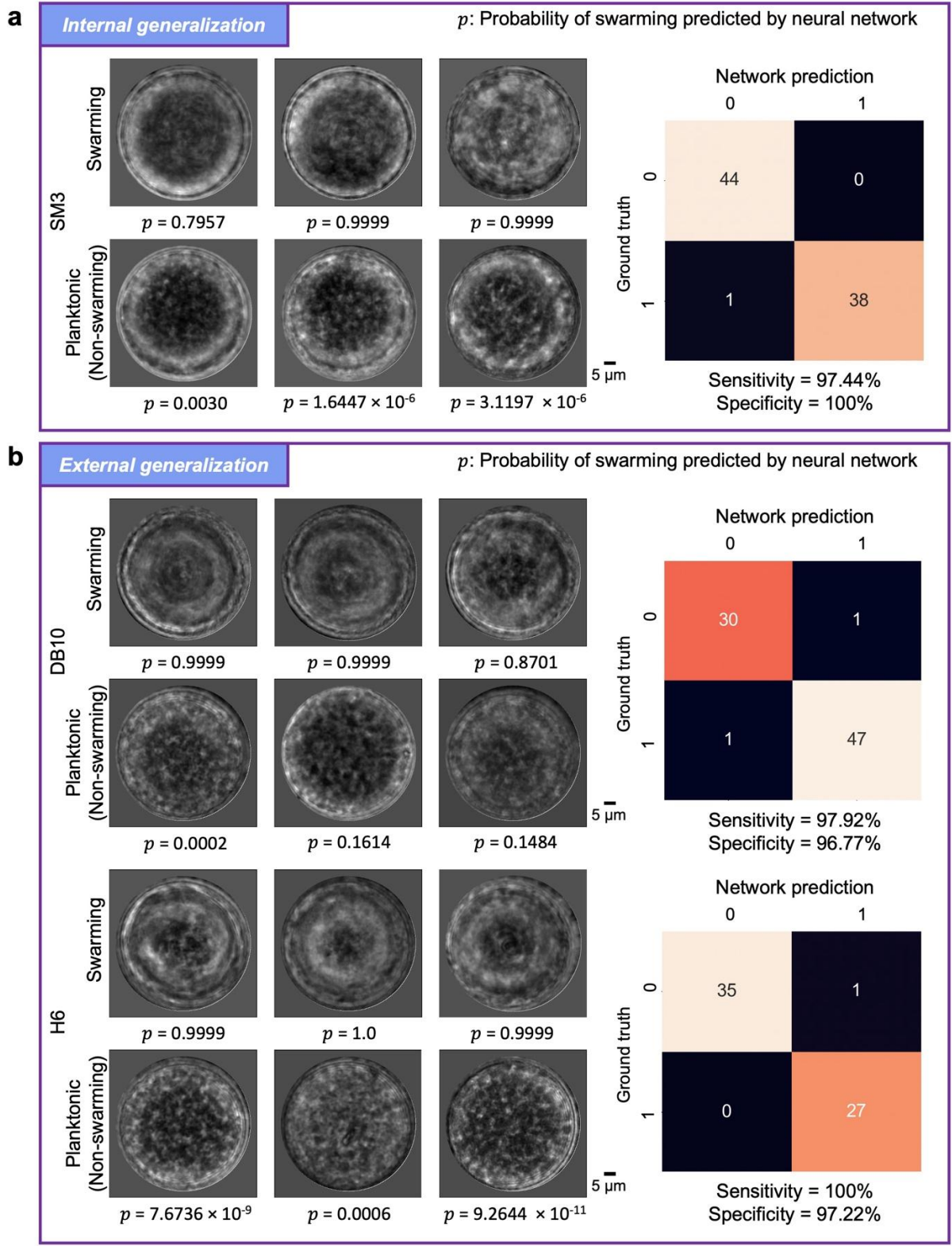

**Figure 3. Swarming detection performance of the deep learning-based classifier in blind testing across SM3, DB10, and H6 bacterial strains. (a)** Three example images for each of the SM3 swarming and swimming/planktonic patterns, along with

the swarming probabilities predicted by our DNN model (left). The confusion matrix tested on a dataset composed of 39 swarming (positive) well images and 44 planktonic (negative) well images, together with the reported sensitivity and specificity are shown on the right to demonstrate the model's internal generalization ability. **(b)** Same as (a), but for the blind testing results of DB10 (48 positive well images and 31 negative well images) and H6 (27 positive well images and 36 negative well images), demonstrating the external generalization capability of the same DNN model.

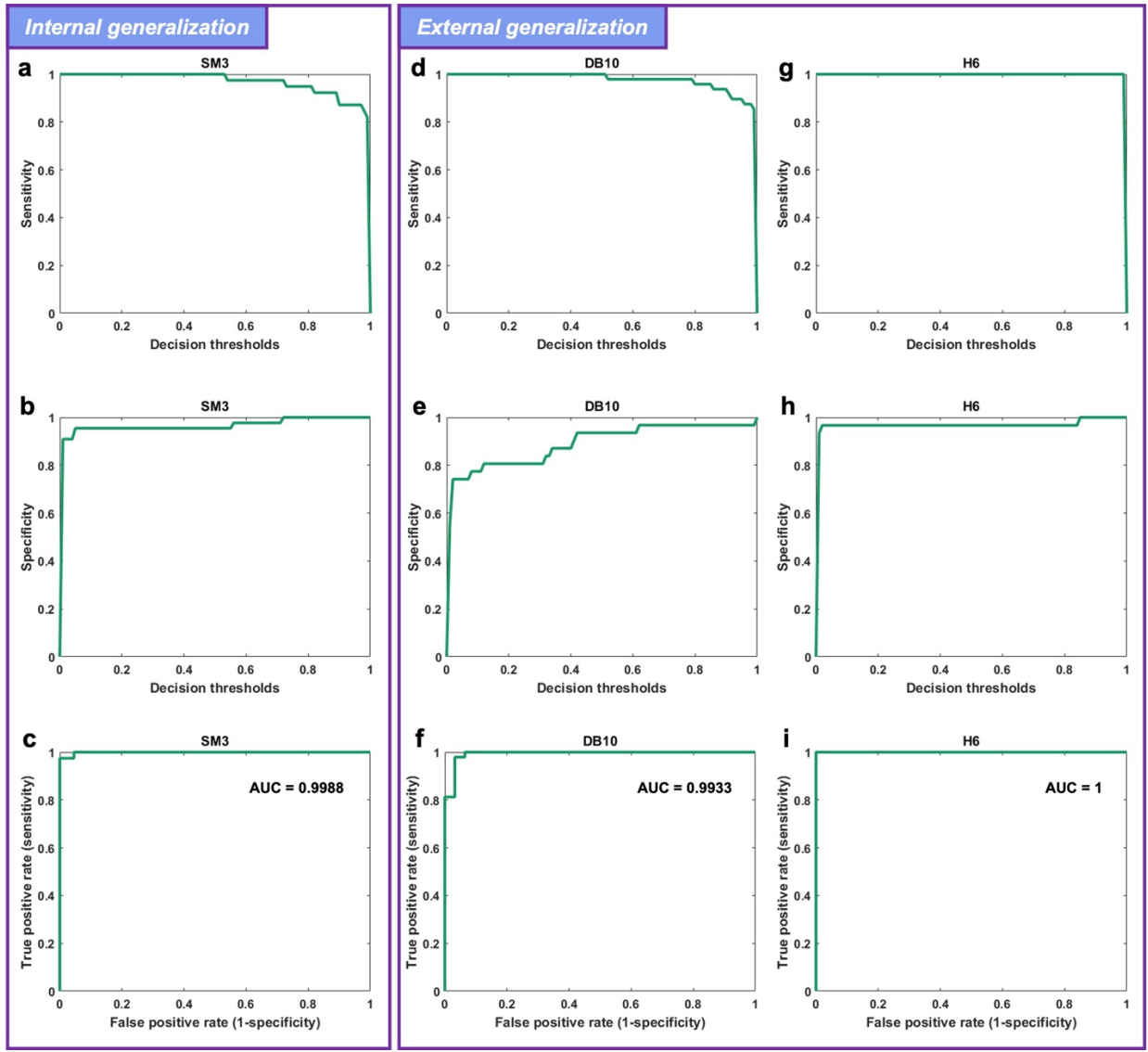

**Figure 4. Sensitivity and specificity curves at varying decision thresholds, alongside the ROC curves for the deep learning-based swarming classifier, in blind testing across SM3, DB10, and H6 bacterial strains. (a)** The sensitivity vs. the decision thresholds for SM3 using the same test dataset in Fig. 3. **(b)** The specificity vs. the decision thresholds for SM3 using the same test dataset in Fig. 3. **(c)** The ROC curve for SM3 using the same test dataset in Fig. 3, with AUC = 0.9988. **(d-f)** Same as (a-c), but for DB10, with AUC = 0.9933. **(g-i)** Same as (a-c) and (d-f), but for H6, with AUC = 1.0.

## Discussion

To underscore the importance of differentiating bacterial swarming behavior, it is instructive to examine the pathological mechanisms of urinary tract and gastrointestinal infections. In addition to the unique role of swarming bacteria as pathogens in UTI, as described previously, bacterial swarming is also closely related to therapeutics for UTI. Specifically, the probiotic, non-swarming *E.coli* 83972, which was endorsed by the European Association of Urology Guidelines in 2015, shows significant preventative (and antibiotic sparing) properties. This strain of bacteria cannot swarm or swim because of poor flagella formation; but can colonize without virulence. A recent report, showed that in individuals with *E.coli* 83972 colonization, who became symptomatic of UTI, the *E.coli* 83972 restored its flagellar machinery and became motile (without a change in other factors). This suggests that motility in UTI can be conditionally pathogenic[34]. Moreover, bacterial swarming colony inhibitors, typically those that affect quorum sensing, are a non-antibiotic strategy used to combat inflammation and virulence[35–43].

Swarming bacteria also play a vital role in gut health. In the normal gut, bacteria downregulate flagella or motility gene synthesis[44], but patients with IBD have an increased number of bacterial swarmers present in colonoscopic aspirates and feces[11,45]. Several pathogens like *salmonella* and *proteus* are implicated in IBD and have swarming properties. These observations directly implicate swarming in pathogenesis, rather than protection against intestinal inflammation. In contrast, our prior work suggests that commensal swarming bacteria could protect against host inflammation and provide a cue to developing personalized probiotics[11]. In the context of IBD, feces from patients showed a significant upregulation of flagella and the motility apparatus as compared to healthy donors[46]. This implies that early swarm flow detection in feces could be developed into diagnostic tests for IBD. More importantly, early isolation of the dominant swarming strain might provide novel probiotics for IBD, with the advantage of this approach being autologous rather than heterologous[45]. These emphasize the necessity to develop efficient and cost-effective methods for detecting swarming bacteria in urine and fecal samples, thereby allowing for rapid diagnosis of infections and associated diseases.

In this work, we developed an AI-based motility test model capable of rapidly and accurately predicting the probability of swarming within each test confinement using a single image with a long integration time of ~0.3 sec. Compared with colonoscopy or fecal microbiome metagenomic analyses, such a non-invasive and economical AI-based motility test would offer several advantages, including reduced discomfort for the patient, cost-effectiveness, rapid result delivery, and user-friendly readouts. Furthermore, apart from the initial sample collection and image capturing, the following detection process is fully automated, requiring minimal human supervision and parameter tuning. This not only saves labor but also ensures an objective and quantitative assessment free from personal bias. Particularly valuable in high-volume settings, our method significantly reduces reliance on human expertise and increases detection throughput compared to human-intervened observations, thereby potentially enabling efficient on-site screening for related diseases. Additionally, our method is adaptable to various application scenarios, allowing for adjusting sensitivity or specificity preferences based on specific needs through flexible decision threshold settings. For instance, in situations where diseases or infections could cause severe consequences, detection sensitivity of our method can be increased by lowering the decision thresholds (with a compromise in specificity). Conversely, in cases where treatments are painful or resource-intensive, detection specificity can be enhanced by setting higher thresholds in our method (with a compromise in sensitivity). Notably, this decision threshold adjustment will be a one-time effort tailored to each specific application scenario, which is inherently different from the intensive parameter tuning for each individual measurement as required by traditional analysis methods. Finally, the single-image-based detection mechanism in our method does not require advanced microscopes with high frame rates. This facilitates the possibility of transferring the AI-based detection algorithm, along with disposable PDMS chips, to portable devices, even smartphones, which could potentially be used to predict real-time swarming probability.

The success of our deep learning-based swarming classifier lies in the incorporation of the attention module to mitigate edge artifacts. These bright, ring-shaped artifacts, present in both positive and negative well images, can interfere with the network's ability to learn useful features, thereby degrading the classifier's performance. To further demonstrate the importance of the attention module used in our workflow, we conducted an ablation study by training another deep learning-based classifier using identical network architectures and training settings, except for the removal of the attention module. The results, presented in Supplementary Figure 1, clearly show the superiority of the model with the attention module, which shows higher sensitivity and specificity for both internal and external generalization tests compared to the model without the attention module. The integration of the attention module not only preserves high model accuracy but also prevents potential data loss that could arise from using a fixed radius for the circular active regions across all input images.

*Future explorations.* Strain-specific motility patterns can be used to distinguish surface adhesion of virulent versus probiotic *E.coli*[47]. These observations may also apply to other bacterial strains where a species may have several sub-strains, some possessing host beneficial, and others host virulent properties. The deep learning-based AI motility detection method demonstrated in our work could potentially be extended to provide predictions for host beneficial or virulent properties based on motility patterns alone. To ensure the prediction accuracy, a comprehensive analysis of swarm and swim motility could be conducted on both soft and hard surfaces, encompassing multiple bacterial strains and different rheological surfaces (e.g., mucins).

A priori knowledge of the strains that are virulent versus beneficial could serve as the ground truth. After the AI model's prediction, beneficial swarming strains could then be curated for various applications such as probiotics or for drug delivery[48], and under external convection methods[49].

Moreover, deciphering swarming type formations under confinement in *mixed* bacterial cultures could be set as future work based on the current deep-learning swarming detection method and then tested combinatorically using multiple swarming and non-swarming species mixed. A goal could be to test the motility detection model in environments that introduce obstacles to flow fields in confinement. For example, it could be tested using actual fecal samples that would have such obstructions (e.g., non-motile food particles). Through these future studies, it might be possible to attain AI-based prediction models for determining swarm-like patterns in complex media like feces or soil. This strategy could eventually be used to monitor host symbiosis or the emergence of a motile pathogen[50], soil bacteria, relationships between soil conditions and plant (crop) health[51], or even climate change via for example, effects on coral bacteria[52–55]. The automated plug-and-play determination of the probability of swarming could be studied in the context of chemo-attractants embedded into the agar to see which environmental or xenobiotic chemicals could trigger acceleration or diminution of swarm behavior. Contextually, these chemicals could serve to either therapeutically inhibit bacterial swarming (e.g., catheter-induced urinary tract pathogens)[52] or aid in the swarming phenotype (e.g., gut-induced inflammation)[56]. Similar approaches have been investigated for individual species of bacteria[57].

*Study limitations*. In our paper, we used selected gram-negative bacterial species isolated from feces (SM3, H6) as well as a laboratory strain (DB10). Indeed, while obligate anaerobes do not swarm[56], there are some gram-positive bacteria that do (e.g., *B. subtilis*) and these types of bacteria have not been examined in detail in the current work. Moreover, the deep learning based swarming detection models in our work were derived using smooth agar surfaces at fixed humidity and ambient temperature conditions, but rheologically different surfaces or agar gel media (e.g., rough agar, mucin, different agar compositions for the swarm plates) have not been examined yet. In the future, these conditions will be thoroughly explored, and using a graded motility library of bacterial mutants (e.g., iCRISPR library[58]) we can potentially adapt the model parameters for a broader context as movement can be altered in subtle ways between strains.

All in all, the plug-and-play automated classification of bacterial motility using a single image with a long integration time of ~0.3 sec is a major step forward from the state of the art. Further studies on mixed bacteria environments in real fecal samples and user-friendly refinements of the current work could potentially lead to the development of a practical application (e.g., for smartphones), capable of capturing images and processing them in real-time, offering a convenient and immediate method for analyzing bacterial motility in various settings.

## Materials and Methods

### Sample preparation and imaging

The bacteria strains used in this study are SM3 (*Enterobacter sp.*, isolated from inflammatory mice by Mani Lab[11]), DB10 (*Serratia marcescens;* lab strain provided by Cornelia Bargmann at Rockefeller University) and H6 (*Citrobacter koseri*, isolated from human colonoscopy aspirates by Mani Lab[11]).

Sample and PDMS preparation procedures were based on the protocols presented in ref[13]. Briefly, bacteria strains were transferred from −80°C glycerol stock to fresh LB (Lysogeny Broth: water solution with 10 g/L tryptone, 5 g/L yeast, and 5 g/L NaCl) and shaken overnight (~16 hours) in a 37°C incubator at 200 rpm. Then, 2 μL overnight bacterial culture was inoculated on the center of an LB agar plate (10 g/L tryptone, 5 g/L yeast, 5 g/L NaCl, and 5 g/L agar; volume = 20 mL/plate) and placed in a 37°C incubator. After the population of bacteria started swarming for 6 hours, a PDMS chip (~1 cm$^2$) was mounted upon the edge of the swarming (positive). Each well on the PDMS chip has a diameter of 74 μm. Similar concentration of freshly grown bacterial culture in LB medium (10 μL) were inoculated on 0.5% LB agar plate, and the PDMS chip was mounted immediately (exhibited planktonic or negative). Once the samples were prepared, they were positioned on a phase-contrast microscope (Olympus CKX41, 20×) for video capture at a frame rate of 29 fps. The imaging field-of-view (FOV) spans approximately 422×353 μm², corresponding to 2448×2048 pixels.

### Image processing and training/validation dataset preparation

After the video acquisition of bacteria exhibiting swarming (positive) or planktonic (negative) status, the training/validation dataset was prepared following the image processing workflow reported in Supplementary Figure 2. First, the centroid coordinates of the individual well/confinement were manually selected for each video. Image sequences centered on these coordinates were then spatially cropped from the entire imaging FOV using a window size of 500×500 pixels (~86×86 μm²) across all time points. To obtain a single image per test well for training/validating the deep learning-based swarming classifier, every 10 consecutive frames of the cropped image sequences were averaged over the time domain, mimicking a 10× longer integration time. The resulting single image was then normalized to have a zero mean and unit variance. Furthermore, to eliminate any interference from background features or noise, the intensity values of the regions outside the well were set to zero. In total, the dataset includes training/validation images from 52 wells across 6 swarming experiments and 38 wells from 10 planktonic experiments. By employing data augmentation over time—varying the start and end points of the 10-frame sequences used for averaging—we generated 1,301 positive and 2,703 negative images that were used for training and validation.

**Other implementation details**

All the image preprocessing and dataset preparations were performed using MATLAB, version R2022b (MathWorks). The codes for training the swarming classification DNN models were developed in Python 3.7.11, utilizing PyTorch 1.10.1. All the network training and testing tasks were carried out on a desktop computer equipped with an Intel Core i9-10920X CPU, 256GB of memory, and an Nvidia GeForce RTX 3090 GPU.

**Network evaluation**

To evaluate the classification performance of our deep learning-based swarming classifier, we used the measurements of sensitivity and specificity, defined as:

$$sensitivity = \frac{TP}{TP + FN} \tag{3}$$

$$specificity = \frac{TN}{TN + FP} \tag{4}$$

where $TP$ means true positives, $TN$ means true negatives, $FP$ means false negatives, and $FN$ means false negatives.

## Supplementary Information

- Network structure and training schedule
- Supplementary Videos 1-2
- Supplementary Figures 1-2